\documentclass[twocolumn,twoside]{article}

\usepackage{PRIMEarxiv}

\usepackage{academicons}
\newcommand{\orcid}[1]{\href{https://orcid.org/#1}{\textsuperscript{\aiOrcid}}}

\usepackage{graphicx}
\usepackage{subcaption}
\usepackage[utf8]{inputenc} 
\usepackage[T1]{fontenc}    
\usepackage{color} 
\usepackage{soul} 
\usepackage{hyperref}       
\hypersetup{
    colorlinks=true,
    citecolor=blue,
    linkcolor=blue,
    filecolor=magenta,      
    urlcolor=blue,
    pdftitle={DETECTAF2-Article},
    pdfpagemode=FullScreen,
    }

\usepackage{url}            
\usepackage{booktabs}       
\usepackage{amsfonts}       
\usepackage{nicefrac}       
\usepackage{microtype}      
\usepackage{lipsum}
\usepackage{fancyhdr}       
\usepackage{graphicx}       
\graphicspath{{media/}}     

\usepackage{multicol}   

\usepackage{fancyhdr}
\title{DETECTA 2.0: RESEARCH INTO NON-INTRUSIVE
METHODOLOGIES SUPPORTED BY INDUSTRY 4.0 ENABLING TECHNOLOGIES FOR
PREDICTIVE AND CYBER-SECURE MAINTENANCE IN INDUSTRIAL SMES
\thanks{Project funded by: Spanish Ministry of Industry, Trade and Tourism through the line of aid to Innovative Business Groupings, in its 2023 call for applications.} 
}

\author{
  \href{https://orcid.org/0000-0003-2165-0144}{\includegraphics[scale=0.06]{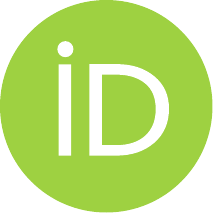}\hspace{1mm}Álvaro Huertas-García}\thanks{Use footnote for providing further information about author (webpage, alternative address)---\emph{not} for acknowledging funding agencies.} \\
  Funditec \\
  Madrid, Spain \\
  \texttt{ahuertas@funditec.es} \\
  \And
  \href{https://orcid.org/0000-0001-5068-3303}{\includegraphics[scale=0.06]{orcid.pdf}\hspace{1mm}Javier Muñoz} \\
  Funditec \\
  Madrid, Spain \\
  \texttt{jmunoz@funditec.es} \\
  \And
  \href{https://orcid.org/0009-0008-1882-9622}{\includegraphics[scale=0.06]{orcid.pdf}\hspace{1mm}Enrique De Miguel Ambite} \\
  Funditec \\
  Madrid, Spain \\
  \texttt{emiguel@funditec.es} \\
  \And
  \href{https://orcid.org/0009-0008-1026-2537}{\includegraphics[scale=0.06]{orcid.pdf}\hspace{1mm}Marcos Avilés Camarmas} \\
  Elliot Cloud \\
  Logroño, Spain \\
  \texttt{marcos.aviles@bosonit.com} \\
  \And
  \href{https://orcid.org/0000-0000-0000-0000}{\includegraphics[scale=0.06]{orcid.pdf}\hspace{1mm}José Félix Ovejero} \\
  Industrias Maxi \\
  Valladolid, Spain \\
  \texttt{josefelix.ovejero@industrias.maxi.es} \\
}

\begin{document}

\twocolumn[ 
  \begin{@twocolumnfalse} 
    \maketitle
    \footnotetext{Project funded by: Spanish Ministry of Industry, Trade and Tourism through the line of aid to Innovative Business Groupings, in its 2023 call for applications.} 
    \bigskip 
  \end{@twocolumnfalse}
]

\begin{abstract}
The integration of predictive maintenance and cybersecurity represents a transformative advancement for small and medium-sized enterprises (SMEs) operating within the Industry 4.0 paradigm. Despite their economic importance, SMEs often face significant challenges in adopting advanced technologies due to resource constraints and knowledge gaps. The DETECTA 2.0 project addresses these hurdles by developing an innovative system that harmonizes real-time anomaly detection, sophisticated analytics, and predictive forecasting capabilities.

The system employs a semi-supervised methodology, combining unsupervised anomaly detection with supervised learning techniques. This approach enables more agile and cost-effective development of AI detection systems, significantly reducing the time required for manual case review. 

At the core lies a Digital Twin interface, providing intuitive real-time visualizations of machine states and detected anomalies. Leveraging cutting-edge AI engines, the system intelligently categorizes anomalies based on observed patterns, differentiating between technical errors and potential cybersecurity incidents. This discernment is fortified by detailed analytics, including certainty levels that enhance alert reliability and minimize false positives.

The predictive engine uses advanced time series algorithms like N-HiTS to forecast future machine utilization trends. This proactive approach optimizes maintenance planning, enhances cybersecurity measures, and minimizes unplanned downtimes despite variable production processes.

With its modular architecture enabling seamless integration across industrial setups and low implementation costs, DETECTA 2.0 presents an attractive solution for SMEs to strengthen their predictive maintenance and cybersecurity strategies.
\end{abstract}


\section{Introduction}
In the context of Industry 4.0, the integration of predictive maintenance (PdM) and cybersecurity for small and medium enterprises (SMEs) represents a crucial advancement towards more reliable and secure operations, since SMEs play an important role in the economy of societies. This integration leverages cutting-edge technologies and methodologies to enhance operational efficiency and safeguard against cyber threats.

Industry 4.0 has transformed predictive maintenance by employing cyber-physical systems (CPS) and the Internet of Things (IoT). These technologies facilitate real-time monitoring and diagnostics, enabling SMEs to preemptively address potential failures and reduce downtime. The use of machine learning models for fault diagnosis further enhances the accuracy and reliability of predictive maintenance systems~\cite{sota_2022,predictive_amintenance_sme_2020}. 

The digital nature of Industry 4.0 exposes SMEs to new cyber risks. To address these, it is vital to integrate robust cybersecurity measures within the predictive maintenance frameworks. This includes using secure cyber-physical systems that provide not only operational efficiencies but also ensure the integrity and confidentiality of industrial data. Advanced algorithms and cybersecurity protocols are essential to protect against data breaches and cyber attacks~\cite{cyber_predictive_maintennace_2020}. 

The DETECTA project, in its second phase, aims to deepen and expand the horizons of Industry 4.0, introducing innovative and non-intrusive methodologies that capitalise on the potential of advanced digital technologies for predictive maintenance and cybersecurity. In an industrial world rapidly moving towards greater connectivity and digitalization, especially in the post-pandemic context, the need to keep production systems not only efficient and productive but also safe and resilient has become imperative.

Funditec, as a central pillar of this endeavour, has positioned itself at the forefront of sustainable technological development. In the course of DETECTA PHASE 1, the foundations of knowledge were laid to successfully address the challenges inherent to anomaly detection and cybersecurity in industrial environments. Through a collaborative approach bringing together clusters, SMEs, and technology centres, significant innovations have been generated that not only encompass the implementation of the Digital Twin and artificial intelligence for predictive maintenance but also better management of cybersecurity in the value chain.

\begin{figure*}
    \centering
\includegraphics[width=\columnwidth]{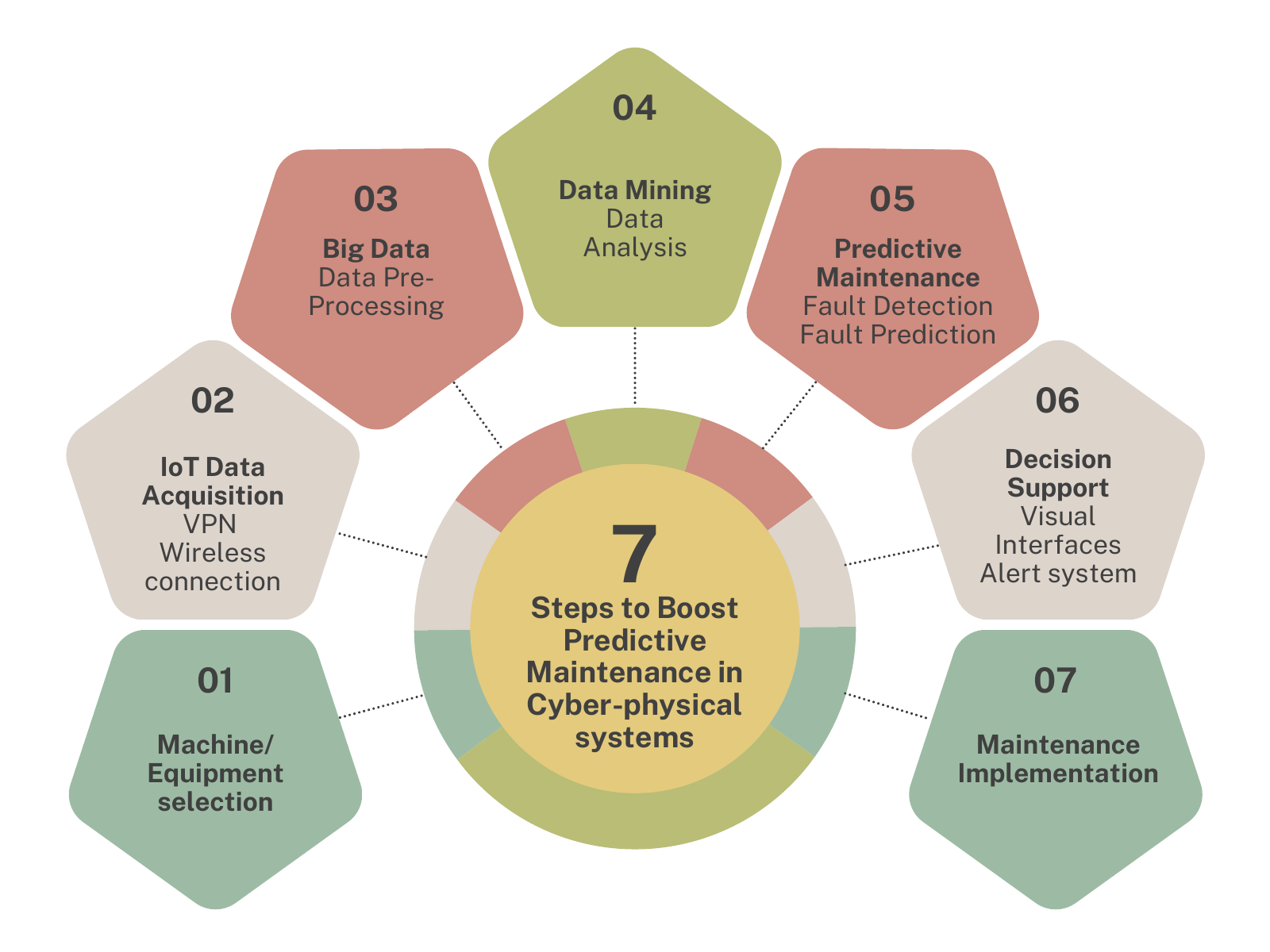}
    \caption{Visualization of system architecture for PdM and cybersecure SMEs in Industry 4.0}
    \label{fig:methodology-overview}
\end{figure*}

\section{Preliminary Analysis: Predictive Maintenance and Cybersecurity in SMEs}

This section outlines the applications of predictive maintenance and cybersecurity in small and medium-sized enterprises (SMEs) in the automotive industry, based on literature, state-of-the-art reviews, and preliminary analyses conducted during DETECTA Phase II across 15 SMEs. This preliminary analysis is done thanks to the survey conducted by the partners from AEI de Ciberseguridad y Tecnologías Avanzadas\footnote{\href{https://www.aeiciberseguridad.es/}{https://www.aeiciberseguridad.es/}} and automotive cluster FACYL\footnote{\href{https://www.facyl.es/}{https://www.facyl.es/}}.

\subsection{Predictive Maintenance (PdM)}

Predictive maintenance (PdM) is becoming increasingly vital for small- and medium-sized enterprises (SMEs) aiming to optimize their industrial operations. PdM leverages continuous monitoring and historical data analysis to forecast potential equipment failures before they occur. This foresight is particularly critical for SMEs where unplanned downtime can severely affect operational efficiency and business reputation~\cite{zhu2024survey}. The ability to anticipate and mitigate these disruptions enhances productivity and reduces maintenance costs by addressing problems proactively. Research, including~\cite{meddaoui_benefits_2023}, highlights the significant efficiencies PdM introduces in manufacturing processes by minimizing operational interruptions. 

Implementing PdM depends on data gathered from sensors embedded within the machinery. These sensors track various operational parameters, offering instant insights into the condition and performance of the equipment. The previous DETECTA project's Phase I~\cite{DETECTAF1} illustrates how SMEs, through AI and advanced analytics integration, can expediently process and analyze large volumes of sensor data to inform decision-making processes.

Despite these advantages, SMEs face substantial challenges in adopting predictive maintenance technologies. Issues such as limited access to necessary technologies, data scarcity, and a lack of expertise can impede effective implementation~\cite{hassankhani_dolatabadi_systematic_2021}.

\subsubsection{PdM Practices in SMEs}

The DETECTA project conducted a market and business survey to assess the current state of technology implementation among SMEs, revealing varied data integration and analysis levels. Although most firms now capture real-time data, particularly those with production processes, the study found that a significant proportion of data still needs to be analyzed due to resource constraints. Specifically, 67\% of companies do not analyze data in real-time, and 40\% typically analyze data post-incident, which undermines the proactive nature of PdM.

Real-time data availability statistics show that only 20\% of companies have comprehensive real-time data across their production processes, while 40\% have partial data, and 27\% lack real-time data altogether. Data collection methods vary, with 20\% of enterprises using company-placed sensors, 33\% utilizing machine-integrated sensors, and the remaining using manual digital or paper-based methods.

These findings indicate a significant gap between data collection and analysis, underscoring the challenges SMEs face in transitioning to data-driven maintenance strategies. Without adequate analysis, predictive maintenance's potential to transform operations and reduce downtime remains largely unexplored.

\subsection{Digital Twin and Cybersecurity}

\subsubsection{Digital Twin Technology in Smart Manufacturing}

Digital Twin technology has emerged as a pivotal element in smart manufacturing, providing dynamic and consistent virtual models of physical entities. These models simulate the characteristics, behavior, life, and performance of their physical counterparts with high fidelity, allowing for real-time operation and prediction. A significant study by the ARC Advisory Group revealed that utilizing digital twins for predictive maintenance can reduce equipment breakdowns by up to 70\% and decrease maintenance costs by approximately 25\%~\cite{digital_twin_report_ARC_2020}.

Recent literature emphasizes the growing application of digital twins in industrial settings for predictive maintenance. These models adapt to changes in the environment and operations, enhancing the accuracy and timeliness of predictions~\cite{liu_digital_2023}. Furthermore, the integration of digital twins with cybersecurity frameworks for industrial control systems has proven beneficial in improving intrusion detection and enabling real-time responses~\cite{digital_twin_intrusion_2022}.

However, digital twins face challenges such as high energy demands and security vulnerabilities in cyber-physical systems. Efforts to enhance the energy efficiency and security features of digital twins are crucial~\cite{digital_twin_proxy_2023}. Moreover, combining artificial intelligence with digital twins has been shown to improve maintenance planning and system reliability significantly~\cite{borangiu_toward_2021}. Systematic reviews advocate for the development of advanced digital twin models to overcome computational and data-related challenges~\cite{VANDINTER2022107008}.

\subsubsection{Cybersecurity Challenges and Measures}

The adoption of digital twin technology introduces specific cybersecurity risks, particularly the threat of cyber-attacks on the virtual representations of physical assets. Effective security measures are essential to safeguard the digital twin and its generated data, thereby preventing unauthorized access and data breaches that could adversely affect the physical systems. These measures include implementing robust access controls, encrypting data, and continuous threat monitoring.

Data quality remains a paramount challenge in maintaining the integrity of digital twins. Ensuring the accuracy, reliability, and consistency of the data used to create and update these models is critical for their effectiveness.

\subsubsection{Cybersecurity Practices in SMEs}
The DETECTA project's preliminary analysis survey provides quantitative insights into the adoption of technologies like digital twins and predictive maintenance among SMEs. The survey indicates that while many SMEs are integrating new technologies, they encounter significant obstacles related to costs, expertise, and time constraints. Cybersecurity is a crucial concern, with considerable investments in training and risk management. However, vulnerabilities in supply chain security remain an issue.
Survey results show varying levels of employee cybersecurity training within SMEs, 27\% rate their training as very high, 46\% as high, 20\% as medium, 7\% as low.

The importance of early anomaly detection is also highlighted, with 60\% of companies considering it critical to distinguish between technical faults and cyberattacks early on. Furthermore, 60\% of surveyed companies have comprehensive risk analyses and proactive protocols covering all employees, but less than half manage cybersecurity actively within their supply chain.

DETECTA project is situated at the vertex of the cutting edge technologies demand aligning with modern frameworks propose the integration of predictive maintenance and cybersecurity into a unified system that encompasses data analytics, machine learning, and IoT. This approach not only predicts equipment failures but also anticipates potential cyber threats, ensuring a comprehensive defense mechanism for industrial operations~\cite{easwaran_cyber_2022}. 

\begin{figure*}[!h]
    \centering
\includegraphics[width=0.8\linewidth]{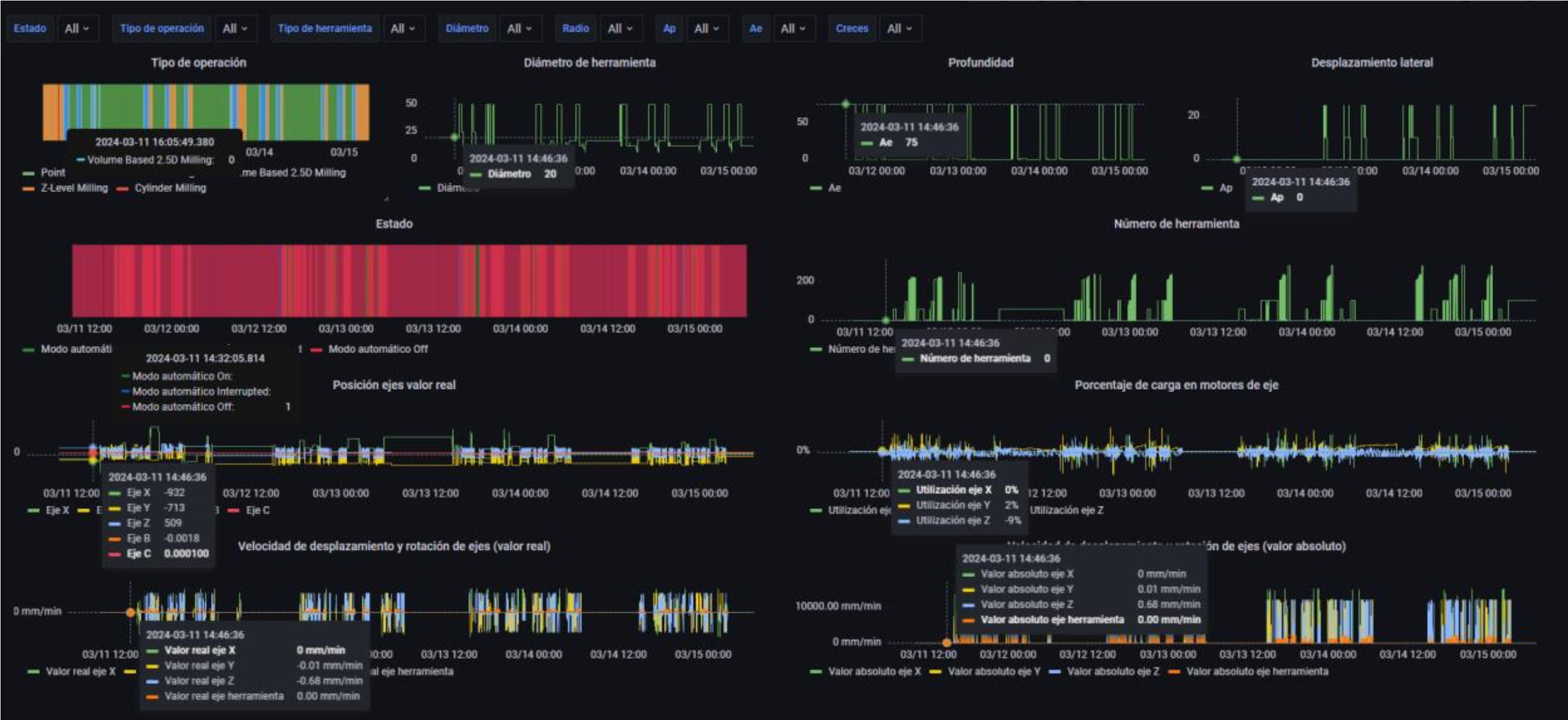}
    \caption{Digital Twin graphical interface created using the open-source software Grafana.}
    \label{fig:DT_UI}
\end{figure*}

\subsection{Research Questions and Contributions}

Following a thorough literature review and preliminary analysis of the current state of Industry 4.0, the DETECTA Phase 2 project builds upon the insights and innovations uncovered in DETECTA Phase 1~\cite{DETECTAF1}. 


The project aims to refine and expand the methodologies and technological tools that enhance the efficiency, safety, and resilience of industrial processes. The project focuses on advanced sensing and monitoring, detailed analysis, and the accurate detection and prediction of production anomalies. The objectives are pursued through the integration of advanced Industry 4.0 technologies, such as big data analytics, machine learning, and deep learning within the IoT domain, with a special focus on strengthening cybersecurity and trust.

DETECTA Phase 2 demonstrates the practical implementation of these methodologies through a case study involving a predictive maintenance solution combined with digital twin technology on a milling machine used by the automotive SME Industrias Maxi\footnote{\href{https://www.industrias-maxi.es/}{https://www.industrias-maxi.es/}} in Spain. The technical research and development undertaken to address the challenges identified in the preliminary stages of the project are discussed in the following sections. This discussion provides insights into the practical implications and benefits of integrating predictive maintenance and digital twin technologies in industrial settings.

\begin{figure*}
    \centering
    \includegraphics[width=1\linewidth]{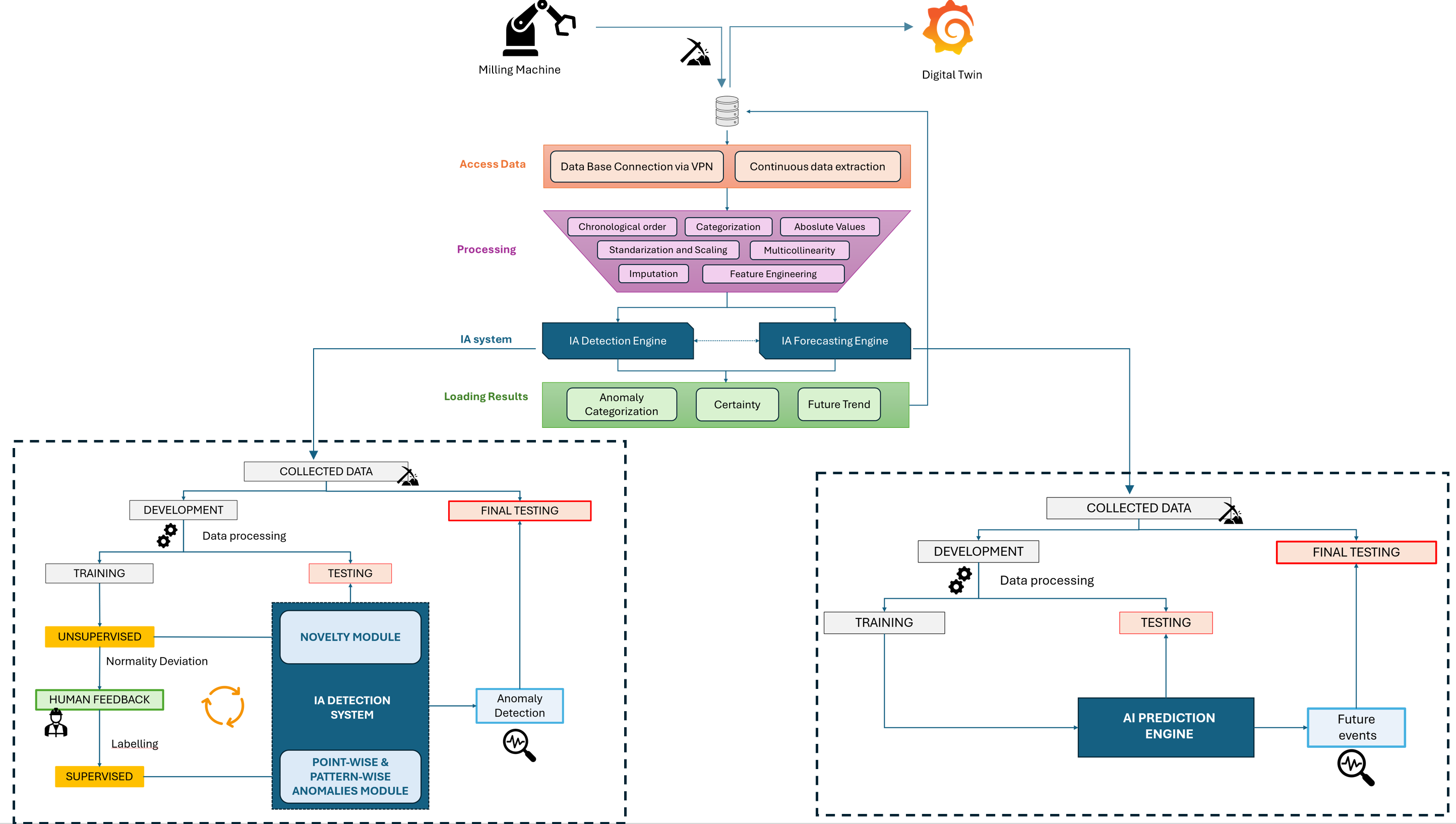}
    \caption{General Schema of the Decision Support and Alert System}
    \label{fig:EDR_method}
\end{figure*}

\section{Case of Study: Anomaly Detection in Milling Machine}

\subsection{IoT Data Acquisition: Digital Twin}


The Heidenhain iTNC 5301 milling machine, designed for five-axis machining with computer numerical control (CNC), has been selected for the development of a digital twin. This machine features extensive sensorization, which is essential for gathering the real-time data necessary for creating an accurate digital twin in an industrial setting.

Data from the milling machine is captured in real-time through IoT agents or connectors utilizing the LSV2 communication protocol. This protocol enables the connectors to interface seamlessly with the machine, allowing efficient data access and transmission without the need for detailed knowledge of the machine's memory locations or data types. The interaction with the milling machine is managed using Python and the PyLSV2 library, providing the flexibility required to handle complex data types and memory positions that are difficult to manage using the manufacturer's standard software development kit (SDK).

The collected data is processed and stored in a PostgreSQL database, ensuring robust data management. Visualization and monitoring of this data are facilitated through Grafana, which offers real-time data visualization capabilities that are well-integrated with PostgreSQL, enhancing the ability to monitor and analyze data effectively.

As shonw in Figure \ref{fig:DT_UI}, the digital twin features a virtual interface that provides real-time monitoring of crucial operational and diagnostic indicators. This interface enables the visualization of various machining processes and allows operators to apply filters to assess different machine states and operations, thereby improving decision-making with enriched data-driven insights. The digital twin monitors several key parameters, including machine state (e.g., automatic, off, or interrupted), operation type, tool characteristics (type, diameter, radius, etc.), axis positioning and movement, and motor usage percentages for each axis.

The functionality of the digital twin includes temporal filtering to customize the monitoring period, selection of specific data filters and variables for targeted analysis, and the capability to display both real-time and historical data, which is crucial for detecting anomalies and optimizing operations. To ensure data security and control, the digital twin operates on an On-Premise server solution within the industrial facility's local network. This arrangement minimizes the risk associated with external data transmission and addresses potential security vulnerabilities effectively.

\subsection{Data Analysis and Data Mining}

In preparation for applying machine learning techniques, an exploratory data analysis (EDA) is conducted on the dataset acquired from the Digital Twin system. This initial phase focuses on identifying key variables that could indicate abnormal behavior in the machine's operations. Among the variables considered, over 43 in number, particular attention is given to factors such as machine state, tool number, position, travel speed, and motor usage for each axis. These variables are integral to the digital twin interface, offering foundational insights for further analysis.

To preserve the integrity of the original data signals, the processing is kept to a minimum. The key steps involved in the data processing phase include:

\begin{itemize}

    \item \textbf{Categorization}: This step involves converting categorical data into a format amenable for analysis, utilizing techniques such as one-hot encoding. This transformation is crucial for preparing the data for subsequent machine learning applications.
    
    \item \textbf{Normalization and Scaling}: Robust scaling techniques are employed to minimize the influence of outliers. This ensures that the analysis is not skewed by extreme values, which can distort the results of machine learning models.
    
    \item \textbf{Handling Multicollinearity}: To ensure model accuracy and efficiency, a threshold is set to eliminate variables with high collinearity. This step helps in reducing redundancy in the dataset, which could otherwise negatively impact the performance of the predictive models.
    
    \item \textbf{Feature Engineering}: The raw data is further refined by deriving additional features such as usage, intensity, speed, and power. These features are crafted to enrich the dataset, providing a more comprehensive basis for detailed analysis.
\end{itemize}

Beyond the variables displayed on the Digital Twin dashboard, the analysis incorporates additional parameters to enhance the model’s diagnostic capabilities. Variables such as utilization, intensity, speed, and power for the axes X, Y, and Z are included. These additional metrics offer deeper insights into the operational state of the machine, enabling more precise anomaly detection and analysis.

\begin{figure*}[htbp]
    \centering
    \begin{subfigure}[b]{1\columnwidth}
        \includegraphics[width=\linewidth]{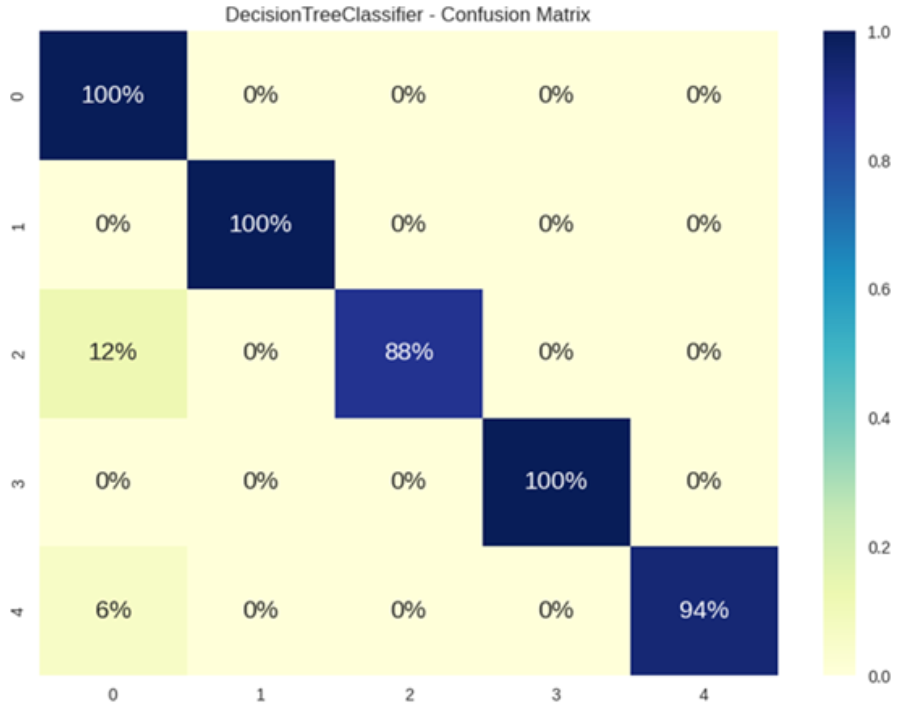}
        \caption{Confusion Matrix of the Decision Tree model selected in the evaluation with unseen data in the detection of anomalies in the predictive maintenance of the milling machine.}
        \label{fig:cm_results_dt}
    \end{subfigure}
    \hfill 
    \begin{subfigure}[b]{1\columnwidth}
        \includegraphics[width=\linewidth]{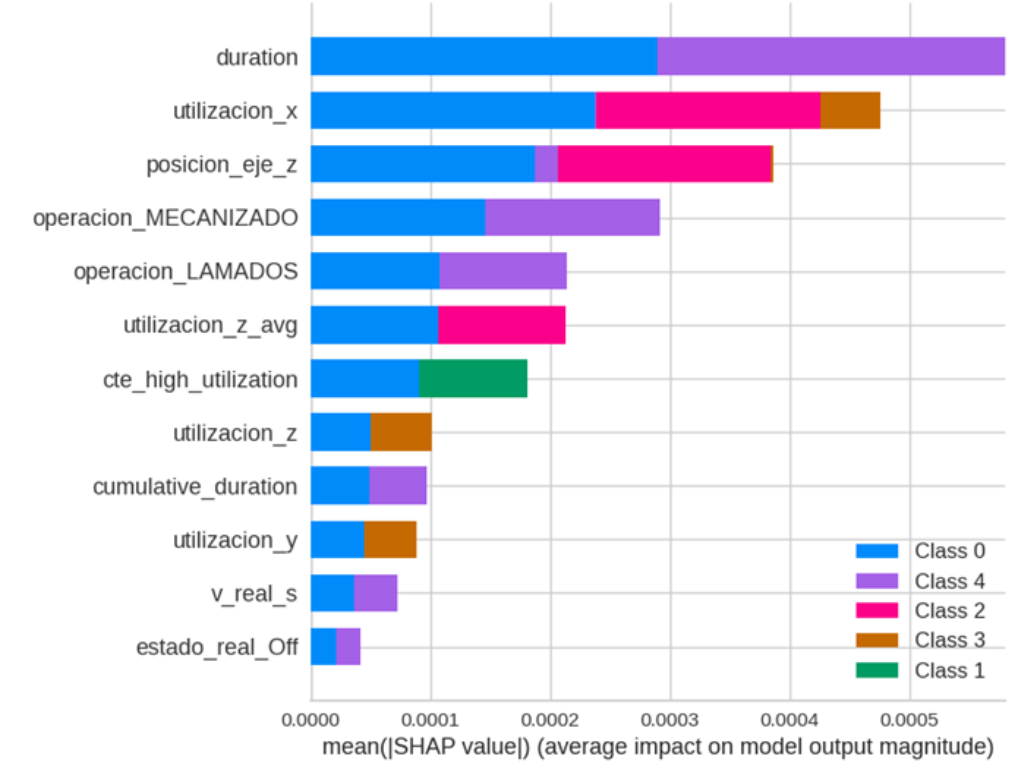}
        \caption{Global SHAP values for the Decision Tree model.}
        \label{fig:shap_global_dt}
    \end{subfigure}
    \caption{Visual analysis of the Decision Tree model using confusion matrix and SHAP values}
    \label{fig:decision_tree_analysis}
\end{figure*}

\subsection{Predictive Maintenance (PdM): Detection and Prediction}

Once the data analysis and preparation pipeline is prepared, we proceed with the predictive maintenance framework for machine fault and cyber incident detection. This framework is developed through the detection of anomalies (see Figure~\ref{fig:methodology-overview}) and the forecasting of the consumption tendency of the milling machine.

\subsubsection{Fault Detection through AI-Powered Anomaly Detection}

Following the data analysis and preparation to maintain signal integrity and enhance analytical effectiveness, an AI-powered anomaly detection engine has been developed to bolster predictive maintenance strategies. This engine employs a semi-supervised approach, combining unsupervised and supervised machine learning techniques to detect unusual patterns that may indicate potential equipment failures~\cite{semisupervised_cite,semisupervised_industry}.

In the initial stage, unsupervised learning methods are utilized to detect new and unforeseen anomalies by identifying data points that deviate significantly from normal patterns. These models include cluster-based models (e.g., Isolation Forest~\cite{islation_forest}), neighbor/distance methods (e.g., Local Outlier Factor~\cite{lof_cite}), projection methods (e.g., PCA~\cite{PCA_cite}), and statistically adjusted methods (e.g., Minimum Covariance Determinant~\cite{hubert_minimum_2010}). Identified potential anomalies are then automatically labeled by propagating the labels using predefined criteria, creating a dataset that trains supervised learning models to recognize similar anomalies in the future, enhancing the predictive maintenance capabilities.

The anomaly detection models are continuously refined and optimized based on incoming data and feedback loops involving human experts who validate and label anomalies. This iterative process, incorporating continuous feedback from operational data and human experts, helps in adapting the predictive maintenance strategies to evolving conditions and improving the accuracy of fault predictions. The AI engine in Phase 2 was capable of identifying several types of operational anomalies critical for predictive maintenance, including a "novelty" category for anomalies that are emergent patterns not previously categorized but detected through the system's enhanced learning capabilities, providing adaptability and continuous learning capabilities to the system.

Once potential anomalies are labeled (manually or automatically), a dataset is created to train supervised learning models to recognize similar anomalies in the future, enhancing the predictive maintenance capabilities. Techniques like hyperparameter optimization using tools like Optuna are employed to improve model accuracy and reduce false positives and negatives.

Various machine learning models were explored for anomaly detection, including linear models (e.g., Logistic Regression), non-linear models (e.g., SVM, QDA), ensemble models (Random Forest Decision Trees), and boosting models (e.g., LightGBM, XGBoost), as well as deep learning models with a particular focus on the performance of the Multilayer Perceptron (MLP). Energy consumption and considerations for Green AI were also taken into account for model selection. Specific metrics such as precision, recall, F1-score, and Matthews correlation coefficient (MCC) were used to evaluate the effectiveness of the fault detection models. The best trade-off model, considering all dimensions of performance and sustainability, was a Decision Tree with 100 single trees and a maximum depth of 12. The performance scores on the unseen test data were 0.96, 0.91 and 0.92 for the F1-macro, Kappa and MCC (see Figure \ref{fig:cm_results_dt}). 

The AI engine was capable of identifying several types of operational anomalies critical for predictive maintenance, including high constant utilization anomalies, elevated utilization with high Z-axis positioning, simultaneous high-speed movements across multiple axes, prolonged and atypical durations in the "Off" state, and "novelty" anomalies which are emergent patterns not previously categorized.

The overall system evaluation demonstrated superiority in terms of linear scalability, reliability as the probability of fault detection is also reported, and maintenance prediction accuracy, reducing the false positive ratio from 40\% in Phase I to 10\% in Phase II. These improvements were supported by enhanced data visualization tools and more robust AI models that could handle larger and more complex datasets.

\begin{figure}
    \centering
    \includegraphics[width=1\columnwidth]{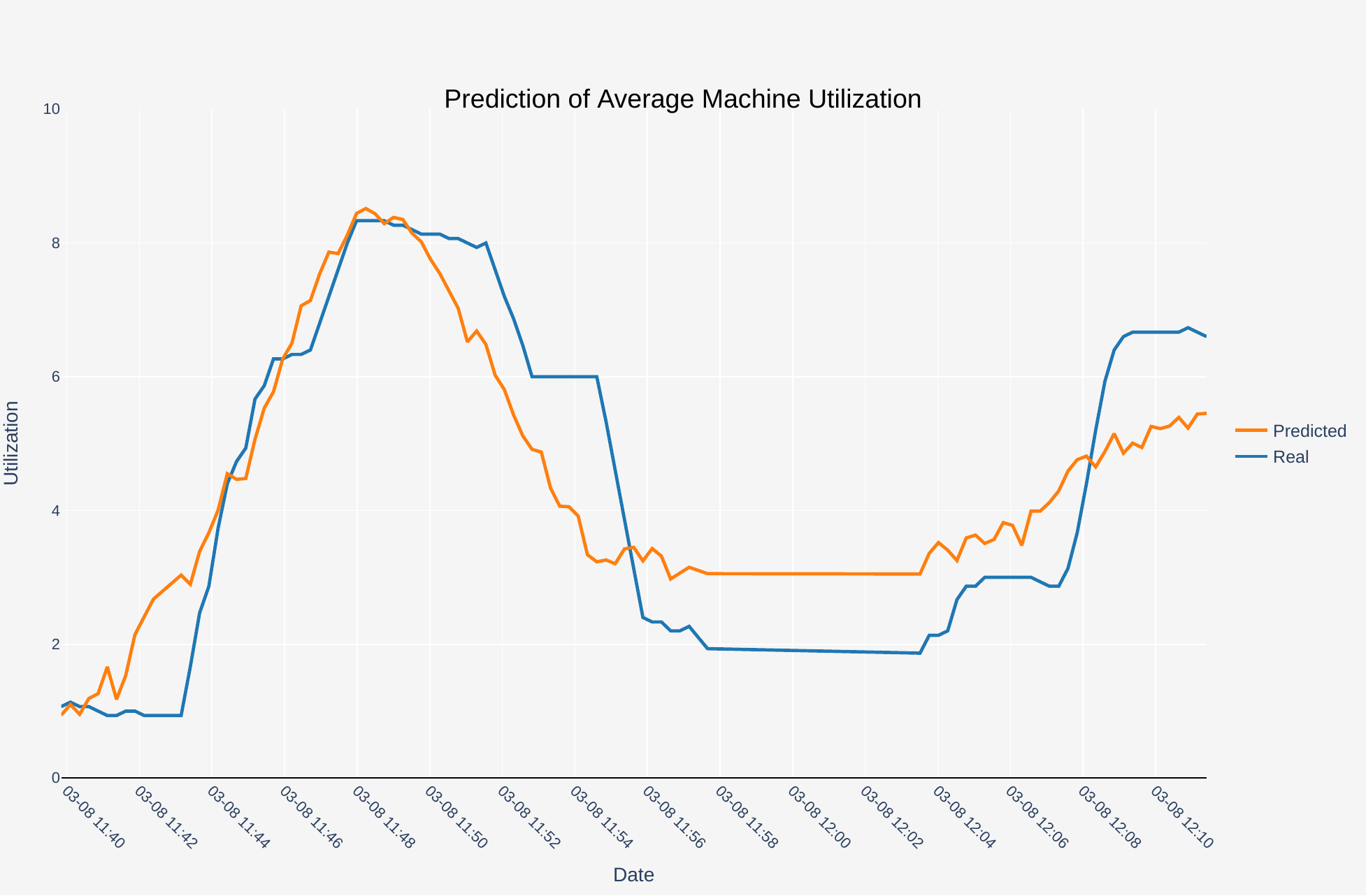}
    \caption{Consumption Forecasting of NHiTS model}
    \label{fig:enter-label}
\end{figure}

\subsection{Consumption Forecasting}

\begin{figure*}[t!]
    \centering
    \begin{subfigure}[b]{1\columnwidth}
        \includegraphics[width= 1\linewidth]{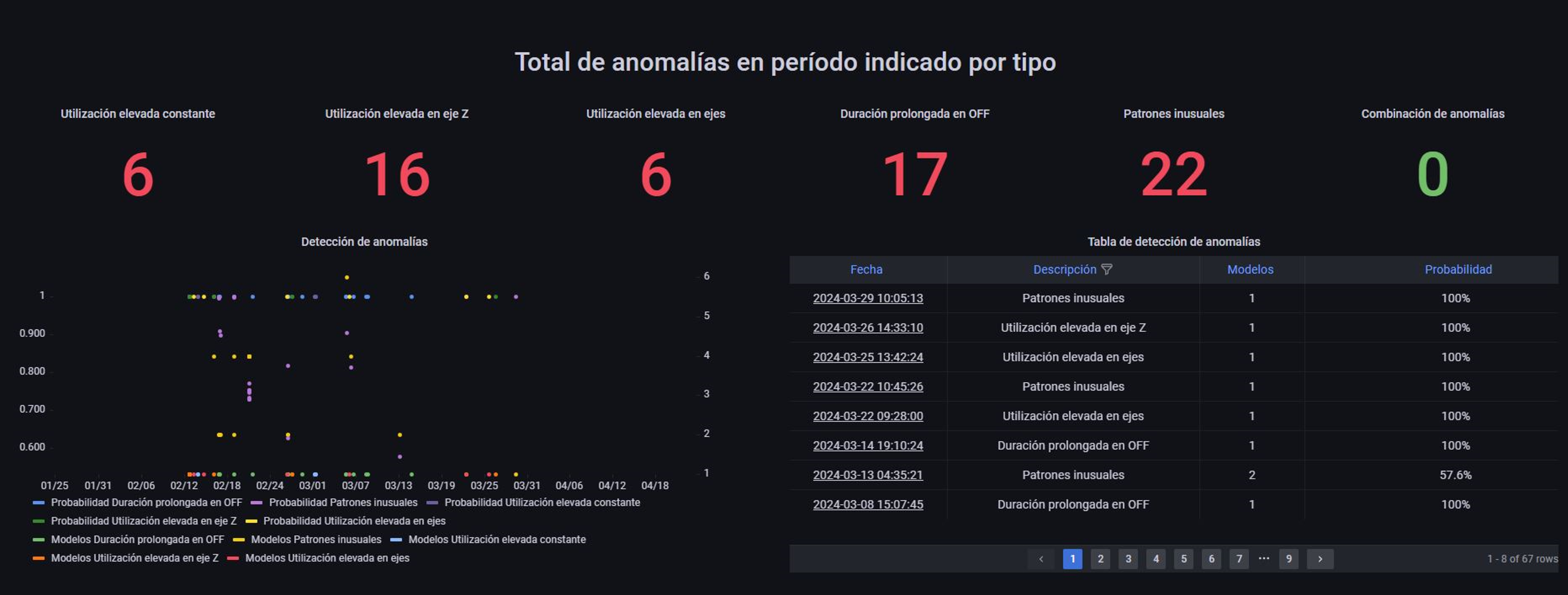}
    \caption{Global vision panel of anomalies detected, with timestamp, type of anomaly and certainty in the Digital Twin}
    \label{fig:edr_panel}
    \end{subfigure}
    \hfill 
    \begin{subfigure}[b]{1\columnwidth}
        \includegraphics[width=1\linewidth]{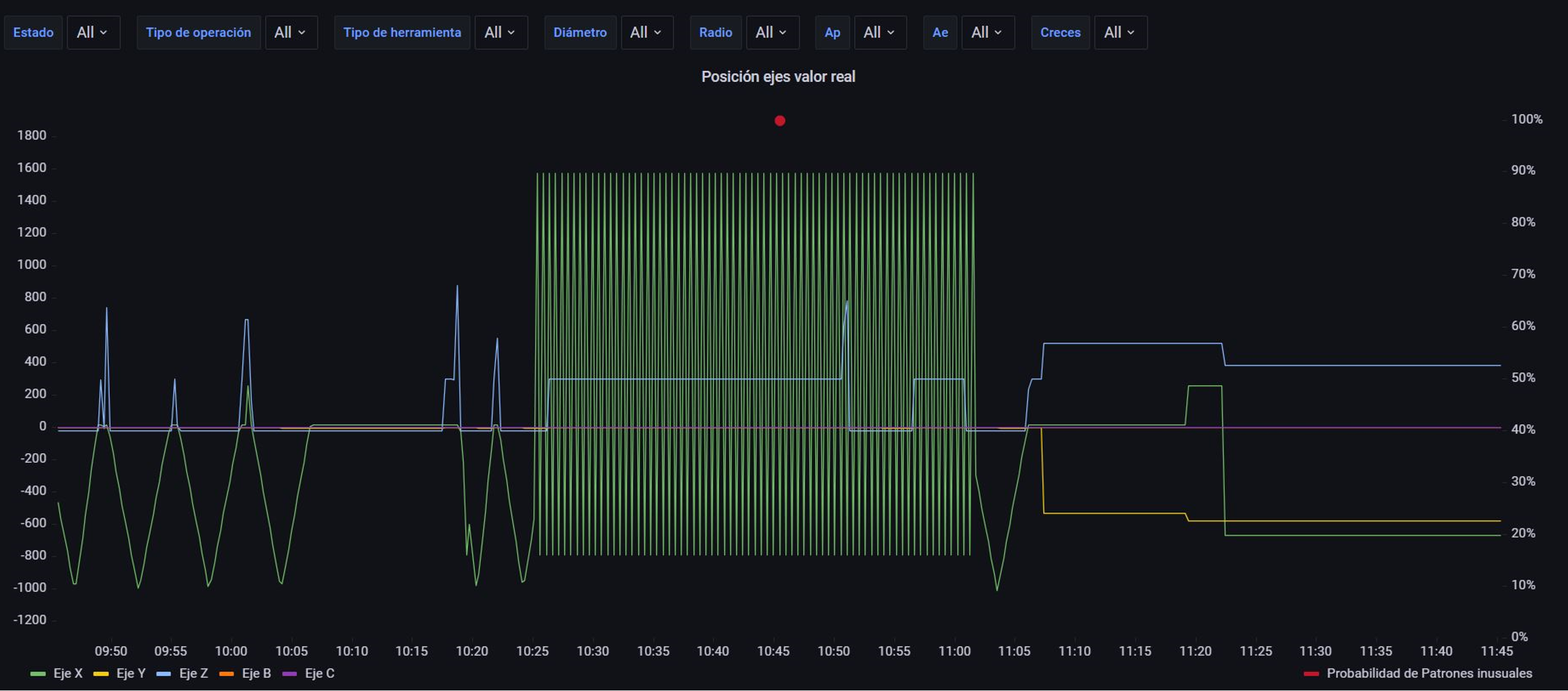}
    \caption{Unusual detected pattern of mechanical failure or cyber incident.}
    \label{fig:edr_example}
    \end{subfigure}
    \caption{Schematic of the Predictive Maintenance methodology overview.}
    \label{fig:EDR-overview}
\end{figure*}

In the era of Industry 4.0, machine data collection and analysis play a pivotal role in optimizing manufacturing processes~\cite{forecasting_industry, wahid_prediction_2022}. The DETECTA 2.0 project applies advanced time series algorithms to predict the utilization of a milling machine, particularly addressing the challenges posed by its variable production process.

Given the unique nature of each part machined, the production process introduces high variability and noise into the data~\cite{forecasting_industry}. This variability in demand means that each part may require different amounts of machine time, complicating the identification of stable patterns. The lack of periodicity, with no evident seasonal or cyclical patterns, further complicates accurate forecasting. Additionally, the machine's utilization depends on external factors such as material supply, machine maintenance, and staff availability, all of which can vary significantly over time.

To tackle these challenges, the N-HiTS (N-Hierarchical Interpolation-based Time Series Forecasting) model is employed~\cite{challu2023nhits}. This model is designed to handle multiple time horizons and capture various temporalities in the data, making it suitable for complex, non-stationary time series.

The N-HiTS model operates through hierarchical decomposition, where the time series is broken down into components at different levels of temporal granularity. This means analyzing data across various time scales (e.g., seconds, minutes, hours) to understand how patterns at one level influence patterns at another. The model uses deep learning-based interpolation, employing neural networks specifically designed to manage these hierarchical decompositions, to learn and predict future values of the time series. This method allows the model to capture seasonal and trend patterns effectively.

A key feature of the N-HiTS model is its ability to perform multi-horizon forecasting, which involves making predictions for multiple future time points simultaneously. This is particularly useful for planning and analysis in scenarios where decisions depend on projections over short, medium, and long terms. The modular and flexible design of N-HiTS allows it to be adapted for different types of time series and specific forecasting requirements.

Optuna~\cite{akiba2019optuna}, a hyperparameter optimization library, is used to fine-tune the model. It automates the search for the best combination of hyperparameters, using advanced search algorithms to improve model performance efficiently.

A predictive window of 100 points, each 15 seconds apart, is chosen to make predictions about the machine's utilization 25 minutes into the future. The model achieves a Mean Absolute Error (MAE) of 1.07, indicating that, on average, the predictions deviate from actual values by this amount. Given the variability of the production process, this is considered a good result. Test results show that the model effectively captures the general trend of the machine’s utilization, though it struggles with sudden spikes or atypical changes.

This predictive capability aids in proactive maintenance and production planning, allowing operators to anticipate future events, optimize workload management, and improve overall operational efficiency. Despite the inherent challenges, the application of sophisticated time series techniques and machine learning provides valuable insights for informed decision-making in dynamic production environments.

\begin{figure*}[!h]
    \centering
    \includegraphics[width=0.6\linewidth]{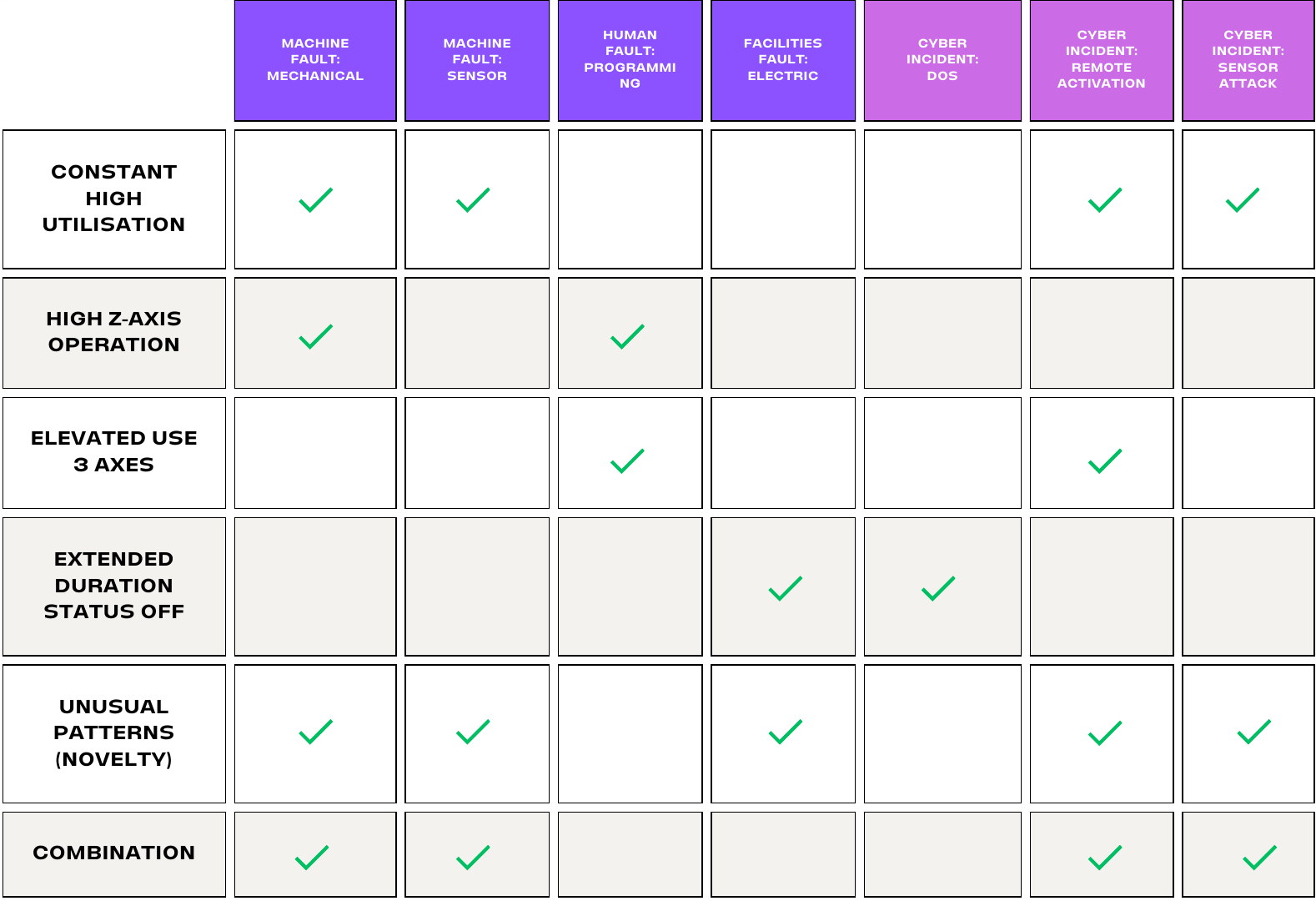}
    \caption{Matrix linking types of incidents and faults to possible causes}
    \label{fig:EDR-matrix}
\end{figure*}

\subsection{Decision Support and Alert System}

As shown in Figure \ref{fig:EDR_method}, the fault detection and forecasting from the AI models are embedded into the digital twin to create the Decision Support and Alert System in DETECTA 2.0. The system leverages real-time anomaly detection and advanced analytics to provide comprehensive monitoring and proactive maintenance for industrial SMEs. 

Anomalies are displayed in real-time using a digital twin interface, as depicted in Figure \ref{fig:EDR-overview}. This interface visualizes key performance parameters and detected anomalies, enhancing situational awareness and enabling prompt responses to potential issues by providing operators with an intuitive view of machine states and alerts.

The AI engines categorize anomalies based on observed patterns, distinguishing between technical errors and cyberattacks. The system identifies various operational anomalies such as prolonged OFF states, unusual usage patterns, and parameter fluctuations that may indicate cybersecurity incidents (see Figure \ref{fig:edr_example}). This categorization helps operators understand and respond appropriately to different anomalies, where Figure \ref{fig:EDR-matrix} shows the matrix of fault detection and possible causes, whether mechanical, human, or cyber threat..

Additionally, the system offers detailed analytics, including the confidence level and the number of AI models corroborating each detection. By measuring the certainty of detections, the system enhances the reliability of alerts, reducing false positives and ensuring timely interventions.

Beyond reactive detection, the system includes a predictive engine that forecasts future utilization trends based on historical and real-time data. This predictive capability aids in proactive maintenance and cybersecurity measures, preparing for potential future anomalies and optimizing machine performance. The integration of predictive analytics ensures effective maintenance planning, minimizing unexpected downtimes and enhancing operational efficiency. The system continuously adapts and refines its algorithms to stay ahead of emerging threats, ensuring robust protection and operational resilience.

\section{Conclusion}

The DETECTA 2.0 project has demonstrated significant advancements in integrating predictive maintenance and cybersecurity for industrial SMEs. By leveraging real-time anomaly detection, advanced analytics, and predictive forecasting, the system offers comprehensive monitoring and proactive maintenance capabilities that enhance operational efficiency and resilience.

The Decision Support and Alert System, embedded within a digital twin interface, gives operators real-time visualizations of machine states and anomalies. This interface, built using Grafana, ensures intuitive situational awareness and enables prompt responses to potential issues. The AI engines' ability to categorize anomalies based on observed patterns and provide detailed analytics, including confidence levels, enhances the reliability of alerts and reduces the likelihood of false positives.
The system's predictive engine, which forecasts future utilization trends based on historical and real-time data, is crucial in proactive maintenance and cybersecurity measures. This capability helps prepare for potential future anomalies, optimize machine performance, and minimize unexpected downtimes. Despite the inherent challenges posed by the variable production process, the use of the N-HiTS model for consumption forecasting has proven effective in capturing general trends and aiding in informed decision-making.
Moreover, the system is highly adaptable to various equipment and industrial environments. Its modular design ensures it can be easily integrated into existing infrastructure without complex setups. This adaptability and low implementation costs make it an ideal solution for SMEs looking to enhance their maintenance and cybersecurity capabilities in the rapidly evolving technological landscape.

Integrating comprehensive monitoring tools and AI-driven analytics also enhances the early detection and response to potential cyberattacks, minimizing their impact on industrial operations. The system's continuous adaptation and refinement of detection algorithms ensure robust protection and operational resilience against emerging threats.

Overall, DETECTA 2.0 exemplifies the successful application of advanced digital technologies in predictive maintenance and cybersecurity, providing a scalable, adaptable, and efficient solution for industrial SMEs. The project's outcomes highlight the importance of integrating cutting-edge AI and machine learning techniques with real-time data processing to enhance operational efficiency, reliability, and security in industrial environments.

\section*{Acknowledgments}
This work was supported by the Spanish Ministry of Industry, Trade and Tourism through the line of aid to Innovative Business Groupings, in its 2023 call for applications under grant AEI-010500-2023-220. The authors would also like to acknowledge the valuable contributions of AEI de Ciberseguridad y Tecnologías Avanzadas and the automotive cluster FACYL in conducting the preliminary analysis and surveys across 15 SMEs during DETECTA Phase II. The DETECTA Phase 2.0 project was developed in collaboration with the SME Industrias Maxi, providing the industrial use case and facilitating the implementation of the predictive maintenance solution on their milling machine.
\bibliographystyle{unsrt}  
\bibliography{references}

\end{document}